\documentclass[a4paper,conference]{IEEEtran}

\ifCLASSINFOpdf
\else
\fi

\usepackage{amsmath}

\usepackage[pdftex]{graphicx}
\usepackage{algorithm}
\usepackage{algpseudocode}
\usepackage{pifont}
\usepackage{mathrsfs,amsmath}   
\usepackage{color, colortbl}

\usepackage[caption=false,font=normalsize,labelfont=sf,textfont=sf]{subfig}

\definecolor{Gray}{gray}{0.9}

\algnewcommand{\Inputs}[1]{%
  \State \textbf{Inputs:}
  \Statex \hspace*{\algorithmicindent}\parbox[t]{.8\linewidth}{\raggedright #1}
}
\algnewcommand{\Initialize}[1]{%
  \State \textbf{Initialize:}
  \Statex \hspace*{\algorithmicindent}\parbox[t]{.8\linewidth}{\raggedright #1}
}

\newcommand{\aggregate}[2]{\underset{#2}{\operatornamewithlimits{#1\ }}}

\begin{document}

\title{Superframes, A Temporal Video Segmentation}

%\author{\IEEEauthorblockN{Hajar Sadeghi Sokeh}
%\IEEEauthorblockA{Faculty of Science, Engineering,\\ and Computing\\
%Kingston University, UK, London\\
%Email: h.sadeghisokeh@kingston.ac.uk}
%\and
%\IEEEauthorblockN{Paolo Remagnino}
%\IEEEauthorblockA{Faculty of Science, Engineering,\\ and Computing\\
%Kingston University, UK, London\\
%Email: p.remagnino@kingston.ac.uk}
%\and
%\IEEEauthorblockN{Vasileios Argyriou}
%\IEEEauthorblockA{Faculty of Science, Engineering,\\ and Computing\\
%Kingston University, UK, London\\
%Email: vasileios.argyriou@kingston.ac.uk}}

\author{
    \IEEEauthorblockN{Hajar Sadeghi Sokeh\IEEEauthorrefmark{1}, Vasileios Argyriou\IEEEauthorrefmark{1}, Dorothy Monekosso\IEEEauthorrefmark{2}, Paolo Remagnino\IEEEauthorrefmark{1}}
    \IEEEauthorblockA{\IEEEauthorrefmark{1}The Robot Vision Team, Faculty of Science, Engineering, and Computing\\
Kingston University, UK, London
    \\\{h.sadeghisokeh, vasileios.argyriou, p.remagnino\}@kingston.ac.uk}
    \IEEEauthorblockA{\IEEEauthorrefmark{2}The Robot Vision Team, Faculty of Science, Engineering, and Computing, Leeds Beckett University, UK, Leeds
    \\\{d.n.monekosso\}@leedsbeckett.ac.uk}
}

% make the title area
\maketitle

\begin{abstract}
The goal of video segmentation is to turn video data into a set of concrete motion clusters that can be easily interpreted as building blocks of the video. There are some works on similar topics like detecting scene cuts in a video, but there is few specific research on clustering video data into the desired number of compact segments. It would be more intuitive, and more efficient, to work with perceptually meaningful entity obtained from a low-level grouping process which we call it `superframe'. This paper presents a new simple and efficient technique to detect superframes of similar content patterns in videos. We calculate the similarity of content-motion to obtain the strength of change between consecutive frames. With the help of existing optical flow technique using deep models, the proposed method is able to perform more accurate motion estimation efficiently. We also propose two criteria for measuring and comparing the performance of different algorithms on various databases. Experimental results on the videos from benchmark databases have demonstrated the effectiveness of the proposed method. 
\end{abstract}

\IEEEpeerreviewmaketitle

\section{Introduction}
In computer vision, many existing algorithms on video analysis use fixed number of frames for processing. For example optical flow or motion estimation techniques~\cite{IMKDB17} and human activity recognition~\cite{6977392,caba2015activitynet}. However, it would be more intuitive, and more efficient, to work with perceptually meaningful entity obtained from a low-level grouping process which we call it `superframe'. 

Similar to superpixels~\cite{Achanta:2012:SSC:2377349.2377556} which are key building blocks of many algorithms and significantly reduce the number of image primitives compared to pixels, superframes also do the same in time domain. They can be used in many different applications such as video segmentation~\cite{Mat_CVPR2010_36247}, video summarization~\cite{conf/cvpr/ChuSJ15}, and video saliency detection~\cite{Tu:2015:VSD:2850604.2850806}. They are also useful in the design of a video database management system~\cite{Aref02avideo} that manages a collection of video data and provides content-based access to users~\cite{IEEEexample:phdurl_Hampapur}. Video data modeling, insertion, storage organization and management, and video data retrieval are among the basic problems that are addressed in a video database management system which can be solved more efficiently by using superframes. By temporal clustering of the video, it is easier to identify the significant segment of the video to achieve better representation, indexing, storage, and retrieval of the video data. 

The main goal of this work is an automatic temporal clustering of a video by analyzing the visual content of the video and partitioning it into a set of units called superframes. This process can also be referred to as video data segmentation. Each segment is defined as a continuous sequence of video frames which have no significant inter-frame difference in terms of their motion contents. Motion is the main criteria we use in this work, therefore we assume all the videos are taken from a single fixed camera.

There is little literature on specifically temporal segmenting of video, and some works on related areas. In this section, we briefly discuss the most relevant techniques to this work: temporal superpixexls, scene cut, and video segmentation.

%\subsection{Temporal superpixels}
The main idea of using superpixels as primitives in image processing was introduced by Ren and Malik in~\cite{RenMalikICCV2003}. Using superpixels instead of raw pixel data is even beneficial for video applications. Although until recently, superpixel algorithms were mainly on the still images, researchers started to apply them to the video sequences. There are some recent works on using the temporal connection between consecutive frames. Reso et al.~\cite{10.1109/ICCV.2013.55} proposed a new method for generating superpixels in a video with temporal consistency. Their approach performs an energy-minimizing clustering using a hybrid clustering strategy for a multi-dimensional feature space. This space is separated into a global color subspace and multiple local spatial subspaces. A sliding window consisting multiple consecutive frames is used which is suitable for processing arbitrarily long video sequences. 

\begin{figure*}[h!]
\centering
\includegraphics[width=6.6in]{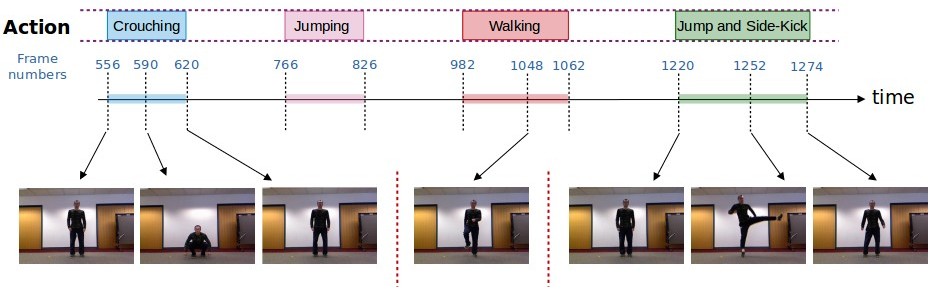}
\caption{An example of superframes in a video from MAD database}
\label{fig:framework}
\end{figure*}

A generative probabilistic model is proposed by Chang et al. in~\cite{chang13tsp} for temporally consistent superpixels in video sequences which uses past and current frames and scales linearly with video length. They have presented a low-level video representation which is very related to a volumetric voxel~\cite{Xu2012EvaluationOS}, but still different in such a way that temporal superpixels are mainly designed for video data, whereas supervoxels are for 3D volumetric data. 

Lee et al. in~\cite{Lee_2017_ICCV} developed a temporal superpixel algorithm based on proximity-weighted patch matching. They estimated superpixel motion vectors by combining the patch matching distances of neighboring superpixels and the target superpixel. Then, they initialized the superpixel label of each pixel in a frame, by mapping the superpixel labels in the previous frames using the motion vectors. Next, they refined the initial superpixels by updating the superpixel labels of boundary pixels iteratively based on a cost function. Finally, they performed some postprocessing including superpixel splitting, merging, and relabeling.

%\subsection{scene cut}
In video indexing, archiving and video communication such as rate control, scene change detection plays an important role. It can be very challenging when scene changes are very small and sometimes other changes like brightness variation may cause false change detection. Many various methods have been proposed for scene change detection. Yi and Ling in~\cite{XiNam2005_sceneChange_ISCAS}, proposed a simple technique to detect sudden and unexpected scene change based on only pixel values without any motion estimation. They first screen out many non-scene change frames and then normalize the rest of the frames using a histogram equalization process.

A fully convolutional neural network has been used for shot boundary detection task in~\cite{DBLP:journals/corr/Gygli17} by Gygli. He considered this work as a binary classification problem to correctly predict if a frame
is part of the same shot as the previous frame or not. He also created a new dataset of synthetic data with one million frames to train this network.

%\subsection{video segmentation}
Video segmentation aims to group perceptually and visually similar video frames into spatio-temporal regions, a method applicable to many higher-level tasks in computer vision such as activity recognition, object tracking, content-based retrieval, and visual enhancement. In~\cite{Mat_CVPR2010_36247}, Grundmann et al. presented a technique for spatio-temporal segmentation of long video sequences. Their work is a generalization of Felzenszwalb and Huttenlocher's~\cite{Felzenszwalb:2004:EGI:981793.981796} graph-based image segmentation technique. They use a hierarchical graph-based algorithm to make an initial over-segmentation of the video volume into relatively small space-time regions. They use optical flow as a region descriptor for graph nodes.  

%baseline
Kotsia et al. in~\cite{DBLP:conf/cvpr/KotsiaA11} proposed using the $3D$ gradient correlation function operating at the frequency domain for action spotting in a video sequence. They used the $3D$ Fourier transform which is invariant to spatiotemporal changes and frame recording. In this work, the estimation of motion relies on the detection of the maximum of the cross-correlation function between two blocks of video frames~\cite{4303017}.

Similar to all these tasks, in this work we propose a simple and efficient motion-based method to segment a video over time into compact episodes. This grouping leads to an increased computational efficiency for subsequent processing steps and allows for more complex algorithms computationally infeasible on frame level. Our algorithm detects major changes in video streams, such as when an action begins and ends.

The only input to our method is a video and a number which shows the desired number of clusters in that video. The output will be the frame numbers which shows the boundaries between the uniform clusters. Visual features like colors and textures per frame are of low importance compare to motion features. This motivates us to employ motion information to detect big changes over the video frames. We use FlowNet-2~\cite{IMKDB17}, a very accurate optical flow estimation with deep networks, to extract the motion between every subsequent frame. We then use both the average and the histogram of flow over video frames and compare our results over a baseline.

\section{The proposed superframe technique}
Our superframe segmentation algorithm detects the boundary between temporal clusters in video frames. The superframe algorithm takes the number of desired clusters, $K$, as input, and generates superframes based on the motion similarity and proximity in the video frames. In other words, the histogram of magnitude is used with the direction of motion per frame and the frame position in the video as features to cluster video frames. Our superframe technique is motivated by SLIC~\cite{Achanta:2012:SSC:2377349.2377556}, a superpixel algorithm for images, which we generalize it for video data (see Figure~\ref{fig:framework}).

\subsection{The proposed algorithm}
Our model segments the video using motion cues. We assume the videos are taken with a fixed camera. Therefore, one can represent the type of motion of the foreground object by computing features from the optical flow. We first apply FlowNet-2~\cite{IMKDB17} to get the flow per video frame, and then we make a histogram of magnitude (HOM) and direction (HOD) of flow per frame to initialize $K$ cluster centers $C_1,\dots,C_K$. Therefore each cluster center has $20$ feature values as $C_k=[HOM_{1..11},HOD_{1..8},fr]^T$ with $k=[1,K]$. $HOM_{1..11}$ indicates `Histogram of Magnitude' and $HOD_{1..8}$ indicates `Histogram of Direction' for $8$ different directions, and $fr$ stands for the frame index in the video.

For a video with $N$ frames, in the initialization step, there are $K$ equally-sized superframes with approximately $N/K$ frames. Since initially, the length of each superframe is $S = N/K$, like~\cite{Achanta:2012:SSC:2377349.2377556}, we safely assume that the search area to find the best place for a cluster center is a $2S \times 2S$ area around each cluster center over the location of video frames.

Following the cluster center initialization, a distance measure is considered to specify each frame belongs to which cluster. We use $D_s$ as a distance measure defined as follows:

Distance between cluster $k$ and frame $i$ is calculated by:
\begin{equation}
\begin{split}
d_c &= \sqrt{\sum{(X_k-X_i)^2}}, \quad X=(x_1\dots x_f) \\
d_s &= fr_k-fr_i \\
D_s &= \sqrt{(\frac{dc}{m})^2+(\frac{ds}{S})^2}
\end{split}
\label{eq_dist}
\end{equation}
where $x$ is a feature value and $X$ is a feature vector of $19$ values per video frame including $11$ values for the histogram of the magnitude of flow and $8$ values for the histogram of the direction of flow. We consider frame location separately as $fr$. $S$ is the interval and $m$ is a measure of compactness of a superframe which regarding the experiment we choose it as $10\%$ of the input number of clusters in this work i.e. $0.1*K$ to make the result comparison easier.

After the initialization of the $K$ cluster centers, we then move each of them to the lowest gradient position in a neighborhood of $3$ frames. The neighbourhood of $3$ frames is chosen arbitrarily but reasonable. This is done to avoid choosing noisy frames. The gradient for frame $i$ is computed as Eq.~\ref{eq_grad}:
\begin{equation}
G(i) = \sqrt{\sum{(X(i+1)-X(i-1))^2}}
\label{eq_grad}
\end{equation}

We associate each frame in the video with the nearest cluster center in the search area of $2S \times 2S$. When all frames are associated with the nearest cluster center, a new cluster center is computed as the average of flow values of all the frames belonging to that cluster. We repeat this process until convergence when the error is less than a threshold.

At the end of this process, we may have few clusters which their length is very short. So, we do a postprocessing step to merge these very small clusters to the closer left or right cluster. The whole algorithm is summarized in Algorithm~\ref{alg1}  

\begin{algorithm}
  \caption{Video superframe clustering algorithm}
  \begin{algorithmic}[1]
    \Inputs{Number of clusters: $K$ }
    \Initialize{m= $0.1*K$  \Comment{compactness}\\ Cluster centers $C_k=[x_1 \dots x_{f}]^T$ at regular step $S$}
    %\Initialize{\strut$w_i^0 \gets 0$, $i=1,\ldots,n$ \\ $S_0 \gets S$}
    \State Perturb cluster centers in a neighborhood, to the lowest gradient position
    \Repeat
      \State $S_t \gets S_{t-1}$
      \For{each cluster center $C_k$}
        \State Assign the best matching frames from a $2S$ neighbourhood around the cluster center according the distance measure (Eq.~\ref{eq_dist}).
      \EndFor
    \State Compute new cluster centers and error $E$ \{$L1$ distance between previous centers and recomputer centers\}
    \Until{$E \leq$ threshold}
    \State Postprocessing to remove very short clusters
  \end{algorithmic}
  \label{alg1}
\end{algorithm}

\subsection{Evaluation criteria for the performance of the algorithm}
Superpixel algorithms are usually assessed using two error metrics for evaluation of segmentation quality: boundary recall and under-segmentation error. Boundary recall measures how good a superframe segmentation adhere to the ground-truth boundaries. Therefore, higher boundary recall describes better adherence to video segment boundaries. Suppose $S=\{S_1,\dots,S_{H} \}$ be a superframe segmentation with $H$ number of clusters and $G=\{G_1,\dots,G_L\}$ be a ground-truth segmentation with $L$ number of clusters. Then, boundary recall $Rec(S,G)$ is defined as follows:

\begin{equation}
Rec(S,G) = \frac{TP(S,G)}{TP(S,G) + FN(S,G)}
\end{equation}
where $TP(S,G)$ is the number of boundary frames in $G$ for which there is a boundary frame in $S$ in range $r$ and $FN(S,G)$ is the number of boundary pixels in $G$ for which there is no boundary frame in $S$ in range $r$. In simple words, Boundary Recall, $Rec$, is the fraction of boundary frames in $G$ which are correctly detected in $S$. In this work, the range $r$, as a tolerance parameter, is set to $0.008$ times the video length in frames based on the experiments.

Under-segmentation error is another error metric which measures the leakage from superframes with respect to the ground truth segmentation. The lower under-segmentation error, the better match between superframes and the ground truth segments. We define under-segmentation error, $UE(S,G)$, as follows:
\begin{equation}
UE(S,G) = \frac{1}{N}\sum_{i=1}^L \bigg( \sum_{j | s_j\cap g_i > \beta} min\{|s_j\cap g_i|,|s_j-g_i|\} \bigg)
\end{equation}
Where $N$ is the number of frames in the video, $L$ is the number of ground truth superframes, $|.|$ indicates the length of a segment in frames and $S_j-G_i=\{x\in S_j|x\notin G_i\}$. By doing some experiments, we found the best number of $\beta$ as an overlap threshold, is $0.25$ of each superframe $s_j$.

\section{Experiments}
This proposed algorithm requires two initialization: $K$ (the number of desired clusters) and `compactness'. In our work, we initialize compactness to $0.1*K$ to make the evaluation of results easier. 

Experiments are carried out using the `MAD'~\footnote{Multimodal Action Database} database and the `UMN' databases. The MAD~\footnote{Videos available from $www.humansensing.cs.cmu.edu/mad/$} database~\cite{database_MAD} is recorded using a Microsoft Kinect sensor in an indoor environment with a total of $40$ video sequences of $20$ subjects. Each subject performs $35$ sequential actions twice and each video is about $4000$--$7000$ frames. We also labelled superframes manually for each video. That shows in which frames there is a big change of motion for the frame, which illustrates the video segments with similar motions, superframes.

UMN~\footnote{Videos available from $mha.cs.umn.edu/proj_events.shtml$} is an unusual crowd activity dataset~\cite{dataset_UMN}, a staged
dataset that depicts sparsely populated areas. Normal crowd activity is observed until a specified point in time where behavior rapidly evolves into an escape scenario where each individual runs out of camera view to simulate panic. The dataset comprises $11$ separate video samples that start by depicting normal behavior before changing to abnormal. The panic scenario is filmed in three different locations, one
indoors and two outdoors. All footage is recorded at a frame rate of $30$ frames per second at a resolution of $640 \times 480$ using a static camera.

Each frame is represented by $20$ features, $11$ values for HOM features, $8$ values for HOD features, and one value for the frame location in the video. HOF features provide us with a normalized histogram at each frame of the video.

\begin{figure}[h!]
\centering
\subfloat[MAD database]{\includegraphics[width=1.5in]{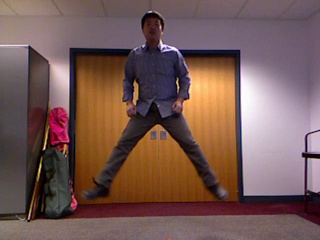}}
\hfil
\subfloat[UMN database]{\includegraphics[width=1.5in]{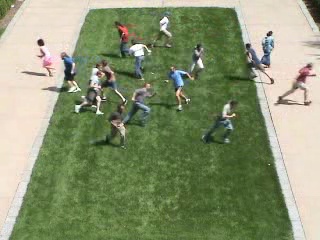}}
\\
\subfloat[Flow/Frame \#326]{\includegraphics[width=1.5in]{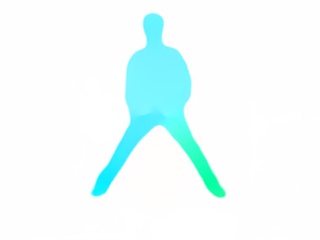}}
\subfloat[Flow/Frame \#257]{\includegraphics[width=1.5in]{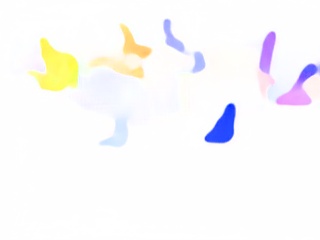}}
\caption{Examples of optical flow results using FlowNet-2 on a frame of a video from MAD and UMN databases.}
\label{fig:opticalflow}
\end{figure}

A dense optical flow~\cite{IMKDB17} is used in this work to detect motion for each video frame. We have employed FlowNet-2 which has an improved quality and speed compared to other optical flow algorithms. With the MAD and UMN databases it took only about $50ms$ and $40ms$ respectively per frame to extract the flow using FlowNet-2. Figure~\ref{fig:opticalflow} illustrates the results of FlowNet-2 on both databases.

To compare our results against, we have used `phase correlation' between two group of frames to determine relative translative movement between them as proposed in~\cite{DBLP:conf/cvpr/KotsiaA11}. This method relies on estimating the maximum of the phase correlation, which is defined as the inverse Fourier transform of the normalized cross-spectrum between two space-time volumes in the video sequence. We call this phase-correlation technique as \textit{PC-clustering} in this work. We sampled every $2$nd frame in each video and chose a sub-section of each frame. We use $240 \times 240$ pixels in the middle of each frame to carry out the phase correlation. Therefore, each space-time volume is considered to be $240\times 240 \times 30$, in which $30$ is the number of frames and shows the temporal length of each volume.

\begin{figure}[h]
\centering
\includegraphics[width=3.0in]{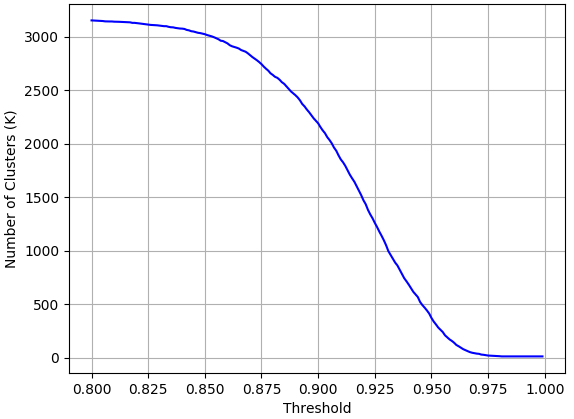}
\caption{The relation between the threshold and the number of clusters in one of the videos of MAD database in PC-clustering technique~\cite{DBLP:conf/cvpr/KotsiaA11}.}
\label{fig:k_thr_FFT}
\end{figure}

\begin{figure}[h]
\centering
\includegraphics[width=3in]{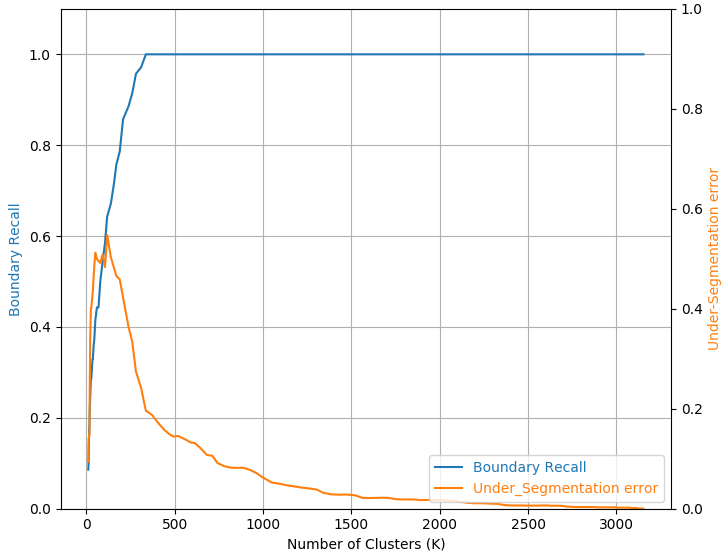}
\caption{Boundary recall and under-segmentation error based on the number of clusters for one of the videos of MAD database for the proposed algorithm.}
\label{fig:rec_UE_K_FFT}
\end{figure}

Given two space-time volumes $v_a$ and $v_b$, we calculate the normalized cross-correlation using Fourier transform and taking the complex conjugate of the second result (shown by $*$) and then the location of the peak using equation~\ref{eq:fft}.  

\begin{equation}
\mathbf{F}_a = \mathcal{F}(v_a), \quad \mathbf{F}_b = \mathcal{F}(v_b), \quad \mathbf{R} = \frac{\mathbf{F}_a \circ \mathbf{F}_b^*}{\mid \mathbf{F}_a \circ \mathbf{F}_b^* \mid}
\end{equation}
where $\circ$ is the element-wise product.

\begin{equation}
r= \mathcal{F}^{-1}\{\mathbf{R}\}, \quad \mathbf{corr} = \aggregate{argmax}{v}\{r\}
\label{eq:fft}
\end{equation}
where $\mathbf{corr}$ is the correlation between two space-time volumes $v_a$ and $v_b$. We calculate this correlation every $2$nd frame of each video and then we conclude that there is a big change in the video when the cross-correlation is less than a threshold (see Figure~\ref{fig:k_thr_FFT}). Figure.~\ref{fig:rec_UE_K_FFT} illustrates how boundary recall and under-segmentation error changes over the number of clusters.

\begin{figure}[h]
\centering
\includegraphics[width=3in]{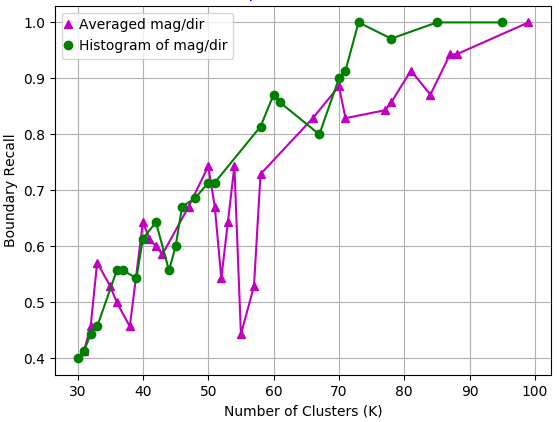}
\caption{Comparison of results when using averaged flow over components of flow and histogram of flow. The former is shown in pink-triangular and the latter is shown in green-circles.}
\label{fig:ave_hist}
\end{figure}

To test the performance of our method on video clustering to superframes, we consider two groups of features: first the averaged value of optical flow over $u$ and $v$, i.e. the $x$ and $y$ components of the optical flow and second the histogram of magnitude and direction of flow. Figure.~\ref{fig:ave_hist} shows the boundary recall with respect to the input number of desired superframes for one of the videos ($sub01\_seq01$) in the MAD dataset. For this video, the number of superframes in the ground truth is $71$ which for $K>70$, HOF features has outstanding improvement on boundary recall over the averaged features.

\begin{figure*}[h]
\centering
\includegraphics[width=6.8in, height=1.8in]{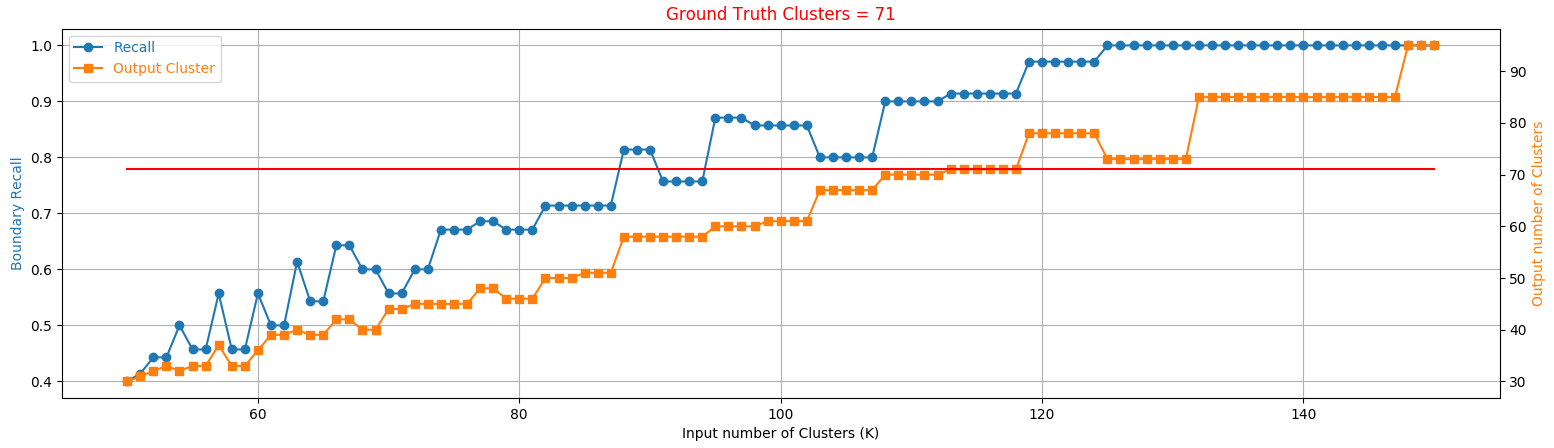}
\caption{Boundary recall and the number of output clusters for a video from MAD database in blue-circle and orange-square, respectively. The number of superframes in the ground truth for this video is $71$ which is shown by a red line.}
\label{fig:HOF_K_Rec}
\end{figure*}

As discussed before, the output number of clusters is usually less than the input number of desired clusters $K$. Since in the postprocessing step some of the clusters may get merged with other clusters as their length is too short. Figure.~\ref{fig:HOF_K_Rec} illustrates the relation between the output number of clusters, $K$ and the boundary recall for a video from the MAD database with $71$ ground-truth clusters. The ground-truth number of clusters is shown using a red horizontal line in the figure. It is shown that when the output number of clusters for this video is $71$ or more, $K$ is more than $115$ and the boundary recall is bigger than $\%91$. 

%Figure~\ref{fig:HOF_K_Rec} illustrates boundary recall for a video from the MAD database with $71$ ground truth superframes. As we stated before, the number of output clusters is more than the input number of clusters, $K$. For example, for $K=115$, the number of clusters detected by the algorithm is $71$, the same as the ground truth, and the boundary recall for this number of clusters is about $\%91$.

\begin{figure}[h]
\centering
\includegraphics[width=3in]{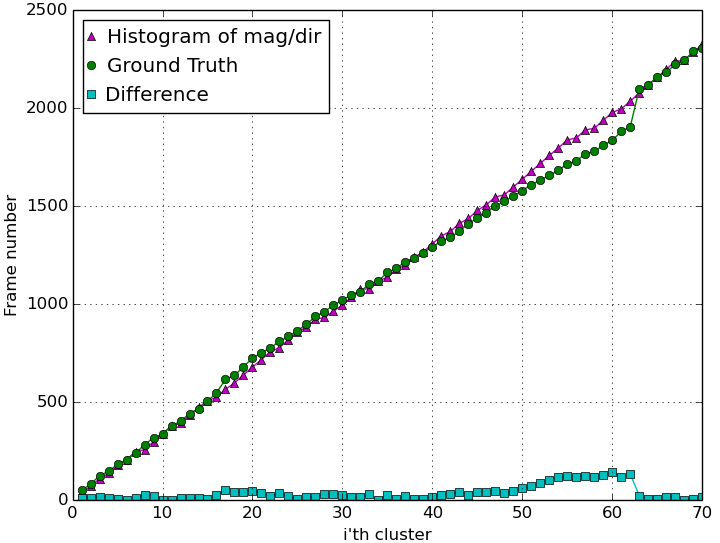}
\caption{Comparison of results when using averaged magnitude/direction and HOF. The number of ground-truth clusters for this video is $71$ clusters and the boundary recall is $0.91$. The averaged flow results are shown in pink-triangular and the HOF is shown in green-circles.}
\label{fig:GT_Res}
\end{figure}

The ground-truth and the result boundaries for a video of MAD database are shown in Figure~\ref{fig:GT_Res}. The difference between them is also shown in this figure which is almost zero except for clusters between $50$ to $62$ between frames $1600$ and $2100$. 

\begin{figure}[h]
\centering
\includegraphics[width=3in]{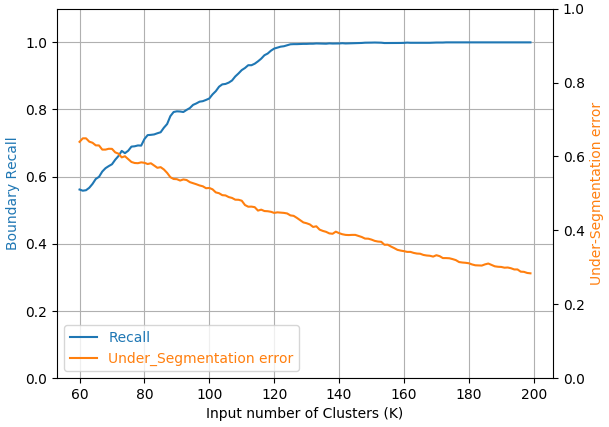}
\caption{Quantitative evaluation of our algorithm by averaging over $40$ videos in MAD dataset.}
\label{fig:rec_usErr}
\end{figure}

As stated before a combination of under-segmentation error and boundary recall is a good evaluation metric for the algorithm. Under-segmentation error accurately evaluates the quality of superframe segmentation by penalizing superframes overlapping with other superframes. Higher boundary recall also indicates less true boundaries are missed. Figure~\ref{fig:rec_usErr} shows the experimental results for both boundary recall and under-segmentation error which are the average values over all $40$ videos in the MAD database. According to this figure, for $K$ more than about $122$, the boundary recall is $\%100$ and the under-segmentation error is less than $\%50$.

A quantitative evaluation of all videos in each dataset is done and the results are illustrated in Table~\ref{tab:compare}. These results are calculated by averaging over all the videos in each dataset when the number of clusters is about $115$. The experimental results show that our model works quite well on both datasets. Regarding this table, the histogram of flow works as a better feature than just averaging the flow or using phase correlation between volumes in the video. There is a $\%22$ improvement of boundary recall for the MAD dataset and $\%8$ for UMN dataset. The boundary recall is quite low for the baseline on UMN dataset, although the segmentation error is still very low. An interesting point in this table is that, there a higher boundary recall for all methods on the MAD dataset, however lower under-segmentation error on UMN datasets. 

\begin{table}[!t]
\caption{Quantitative evaluation for video clustering}
\label{table_example}
\centering
\begin{tabular}{c||c|c||c|c}
\hline
 & \multicolumn{2}{c||}{MAD database} & \multicolumn{2}{c}{UMN database}\\
 \cline{2-5}
 & UE & Rec & UE & Rec \\
 \hline\hline
 \rowcolor{Gray}
PC-clustering~\cite{DBLP:conf/cvpr/KotsiaA11} & $0.54$ & $0.64$ & $0.27$ & $0.25$\\
Averaged flow & $0.52$ & $0.77$ & $0.19$ & $0.63$\\
\rowcolor{Gray}
Histogram of flow (proposed) & $\mathbf{0.31}$ & $\mathbf{0.99}$ & $\mathbf{0.17}$ & $\mathbf{0.71}$\\
\hline

\end{tabular}
\label{tab:compare}
\end{table}

\section{Conclusion}
In this paper, we proposed using Histogram of Optical Flow (HOF) features to cluster video frames. These features are independent of the scale of the moving objects. Using these atomic video segments to speed up later-stage visual processing, has been recently used in some works. The number of desired clusters, $K$, is the input to our algorithm. This parameter is very important and may cause over/under-segmenting the video.

Our superframe method divides a video into superframes by minimizing a cost function which is distance-based and makes each superframe belong to a single motion pattern and not overlap with others. We used FlowNet-2 to extract motion information per video frame to describe video sequences and quantitatively evaluated out method over two databases, MAD and UMU. 

One interesting trajectory for the future work is to estimate the saliency score for each of these superframes. This helps us to rank different episodes of a video by their saliency scores and detect the abnormalities in videos as the most salient segment.

This work was supported in part by the European Unions Horizon 2020
Programme for Research and Innovation Actions within IoT (2016): Large
Scale Pilots: Wearables for smart ecosystem (MONICA).

\bibliography{mybib}{}
\bibliographystyle{IEEEtran}

\end{document}